\documentclass[11pt,a4paper]{article}
\usepackage[hyperref]{acl2020}
\usepackage{times}
\usepackage{latexsym}

\usepackage{microtype}

\usepackage{url}
\usepackage{booktabs}
\usepackage{scrextend}
\usepackage{graphicx}
\usepackage{textpos}
\usepackage{xcolor}
\usepackage{tikz}
\usepackage[ruled]{algorithm}
\usepackage[noend]{algpseudocode}
\usepackage[normalem]{ulem}
\usepackage{colortbl}
\usepackage{microtype}
\usepackage{siunitx}
\usepackage{amsmath}
\usepackage{amssymb} % \mathbb
\usepackage[shortlabels]{enumitem}% noitemsep
\usepackage{multirow}
\usepackage{lscape}
\usepackage{rotating}

\aclfinalcopy % Uncomment this line for the final submission
\def\aclpaperid{1471} %  Enter the acl Paper ID here

\title{Empowering Active Learning to \\ Jointly Optimize System and User Demands}

\author{Ji-Ung Lee \\ \And
  Christian M. Meyer \\
    \texttt{\{lastname\}@ukp.informatik.tu-darmstadt.de} \\
  Ubiquitous Knowledge Processing (UKP) Lab \\
  Computer Science Department \\
  Technische Universit\"at Darmstadt, Germany \\
  {\url{https://www.ukp.tu-darmstadt.de}} \And
  Iryna Gurevych \\
  }

\date{}

\begin{document}
\maketitle
\begin{abstract}
Existing approaches to active learning maximize the system performance by sampling unlabeled instances for annotation that yield the most efficient training.
However, when active learning is integrated with an end-user application, this can lead to frustration for participating users, as they spend time labeling instances that they would not otherwise be interested in reading.
In this paper, we propose a new active learning approach that jointly optimizes the seemingly counteracting objectives of the active learning system (training efficiently) and the user (receiving useful instances). 
We study our approach in an educational application, which particularly benefits from this technique as the system needs to rapidly learn to predict the appropriateness of an exercise to a particular user, while the users should receive only exercises that match their skills.
We evaluate multiple learning strategies and user types with data from real users and find that our joint approach better satisfies both objectives when alternative methods lead to many unsuitable exercises for end users.\footnote{Our code and simulated learner models are available on Github: \url{https://github.com/UKPLab/acl2020-empowering-active-learning}}
\end{abstract}

\section{Introduction}
\label{sec:introduction}
State-of-the-art machine learning approaches require huge amounts of training data.
But for many NLP applications, there is little to no training data available. % yet only %at all.  
\emph{Interactive NLP systems} are a viable solution to alleviate the cost of creating large training datasets before a new application can be used.
Such systems start with no or few labeled instances and acquire additional training data based on user feedback for their predictions.
\emph{Active learning} \citep{Settles2012} is a frequently used technique to quickly maximize the prediction performance, as the system acquires user feedback in each iteration for those instances that likely yield the highest performance improvement (e.g., because the system is yet uncertain about them).
Active learning has been shown to reduce the amount of user feedback required while improving system performance for interactive NLP systems \citep{Avinesh2017, Gao2018} and to reduce the annotation costs in crowdsourcing scenarios \citep{Fang2014}.
However, outside the typical annotation setup, it can be boring or frustrating for users to provide feedback on ill-predicted instances that hardly solve their needs.
Consider a newly launched web application for learning a foreign language, which aims at suggesting exercises that match the user's proficiency according to Vygotsky's \emph{Zone of proximal development} \citep{Vygotsky78}.
The underlying machine learning system starts without any data, but employs active learning to select an exercise the system cannot confidently predict. 
Then, it adjusts its model interactively based on the user's feedback.
While the system is still uncertain, the users often receive inappropriate  (e.g., too hard or too easy) exercises.
Thus, they get the impression that the system does not work properly, which is especially harmful during the inception phase of an application, as the community opinion largely defines its success.

\begin{figure*}
  \centering
  \includegraphics[width=\linewidth,trim={1.7cm 7.8cm 1cm 6cm},clip]{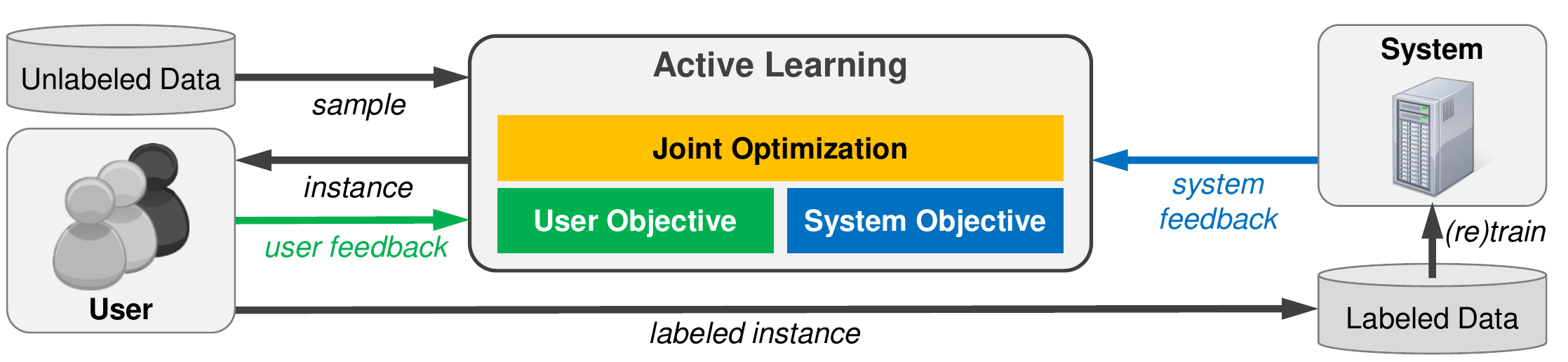}
  \caption{Overview of our interactive approach. We go beyond previous work on optimizing the system objective (blue) by modeling the user objective (green) and jointly optimizing these seemingly counteracting goals (gold).}
  \label{fig:approach}
\end{figure*}

In this paper, we distinguish the \emph{system objective} of maximizing the prediction performance with minimal labeled instances and the \emph{user objective} of providing useful instances for the user's current needs.
For the first time, we propose an active learning approach that jointly optimizes these seemingly counteracting objectives and thus trades off the demands of system and user.

The users of educational applications can particularly benefit from this, as they can learn most if they receive appropriate learning material while the underlying system requires considerable training to reach acceptable performance.
We employ our new approach in a language learning platform for C-tests (i.e., cloze tests, in which the second half of every second word is replaced by a gap).
Our system successfully learns how to predict the difficulty of a C-test gap (system objective) and how to provide a C-test that is neither too easy 
for the current user, which would cause boredom, nor too hard, which would create frustration (user objective).
Predicting the difficulty of an exercise and correspondingly selecting exercises that match a user's proficiency are important steps towards self-directed language learning and massive open online courses (MOOCs) on language learning.
Though we focus on this educational use case in this paper, our approach may also yield new insights for other problems that suffer from seemingly counteracting system and user objectives, for example, interactively trained recommender systems for books, movies, or restaurants.

\section{Related Work}
\label{sec:relwork}
\paragraph{Active learning.}
Active learning aims to reduce the amount of training data by intelligently sampling instances that benefit the model most \citep{Settles2012}.
A distinct characteristic of active learning is that labels for sampled instances are unknown and provided by an oracle after sampling.
Various works investigate the use of active learning for crowdsourcing, where the oracles (i.e., the crowdworkers) may provide noisy labels \citep{Snow2008,Laws2011}.
Within the educational domain, active learning research is scarce.%
\footnote{Note, that in education, active learning often refers to a teaching paradigm which is unrelated to active learning in machine learning.}
One example is the work by \citet{Rastogi2018}, who propose a threshold-based sampling strategy utilizing the prediction probability and achieve a considerable speed-up without any significant performance drop.
\citet{Hastings2018} find that active learning can be used to efficiently train a system for providing feedback on student essays using teachers as oracles.
\Citet{Horbach16} report mixed results for employing active learning in short-answer grading.
While all of these works focus on improvements of the proposed system, users only benefit after training.
In contrast, our work explicitly models the user objective, such that users already benefit while labeling training instances. 

\paragraph{Adaptive learning.}
Many systems provide user adaptation, and research has shifted from pre-defined sets of rules for adaptation to data-driven approaches.
Several works investigate adaptive methods to provide exercises which are neither too hard nor too boring.
For instance, \citet{Missura2011} model learning in a game-theoretic sense where the goal is to adjust the difficulty to neither being too easy nor too hard.
Other works investigate adaptation in the context of testing \citep{Zheng2015, Wang2016, Chaimongkol2016} and propose methods for an adaptive selection of appropriate tests for better assessing a student's proficiency.
In a large survey, \citet{Truong2016} discusses how to integrate different learning styles, modeling categorical student behavior, into an adaptive learning environment and emphasizes the need for more sophisticated methods.

Despite much research in adaptive and active learning, none of the previous works consider jointly modeling and optimizing both the system and user objectives which may retain a user's motivation and keep them from leaving the platform due to boredom or frustration.

\section{Approach}
\label{sec:approach}
Figure~\ref{fig:approach} shows our proposed interactive learning setup. 
The \emph{active learning} component iteratively samples instances from a pool of unlabeled data and asks the user for a label that can be used to train the machine learning system. 
Previous work on active learning focused on optimizing the \emph{system objective} (blue).
That is, only the system provides feedback to the active learning component (e.g., how certain it is about the predicted label of an instance).
In our work, we first model the \emph{user objective} (green) and propose sampling strategies that maximize the user satisfaction based on the user's feedback (e.g., the user's label for an instance).
Finally, we study our novel \emph{joint optimization} strategies (gold) that trade off the demands of the system and the users.
Whereas we distinguish between the user's feedback (exercise-level) and labeled instances (gap-level) in our work, our proposed approach can easily be adapted to more specific cases where the (implicit) user feedback and the provided label are the same.\footnote{Note, that from a single answer which is either correct or wrong, we cannot deduce a fine-grained gap label. To obtain these in a real-world setting, one either may assume querying groups of users or asking them for an explicit label.}

In the remainder of this section, we introduce \emph{sampling strategies} that select which instance should be presented to the user next. 
We use the following notation:
Let $\mathcal{X}$ be the pool of unlabeled instances.
In every iteration of the application (e.g., when a user requests a new exercise), the sampling strategy $s(v)$ returns an instance $x \in \mathcal{X}$ for user $v$.
The user then provides a label $y$ for instance $x$, potentially with additional feedback on the user's satisfaction.
The active learning component finally removes $x$ from its pool $\mathcal{X}$ and adds $(x, y)$ to the set of labeled instances, before the system is retrained with the increased labeled training set.

The simplest sampling strategy that we use as a baseline is \emph{random sampling} $s_\mathrm{rand}(v)$, which selects an $x \in \mathcal{X}$ uniformly at random, regardless of the user.
In the following subsections, we discuss more advanced strategies that optimize the system or user objective as well as our new joint optimization strategies.

\subsection{System optimization}
To optimize the system objective, we consider \emph{uncertainty sampling} \citep{Lewis1994}.
Uncertainty sampling assumes that instances for which the model is least certain during prediction provide the most information for the model once their labels are known.
The sampled instance is thus
\begin{equation}\label{eq-uncertainty-sampling}
  s_\mathrm{unc}(v) = \arg\max_{x \in \mathcal{X}} \: U(x)
\end{equation}
where $U\colon x \mapsto [0,1]$ 
returns the uncertainty of predicting a label for instance $x$.
Like random sampling, $s_\mathrm{unc}(v)$ is independent of the current user $v$.
A model's uncertainty can be measured in multiple different ways, for example, by the prediction probability of the predicted label \citep{Lewis1994}, as the difference in probabilities between the first and second most probable labels \citep{Scheffer2001}, and based on the Shannon entropy \citep{Shannon1948} that considers all possible labels \citep{Settles2008}.
We instantiate $U$ for our educational application in section~\ref{sec:instantiation}.

\subsection{User optimization}
The objective of users is to receive instances that meet their demands.
We therefore define a new \emph{user-oriented sampling} strategy as
\begin{equation}
  s_\mathrm{usr}(v) = \arg\max_{x \in \mathcal{X}} \: A(x, v)
\end{equation}
where $A\colon (x,v) \mapsto [0,1]$ returns the degree of appropriateness of instance $x$ for the user $v$.
In our educational application, we consider an exercise appropriate if it is neither too easy nor too difficult, as this maximizes the user's learning gain.
To quantify $A$, we measure the error between the predicted label $f(x)$ and the user's demand $\phi(v)$ as 
\begin{equation}
  A(x, v) = 1 - \mathrm{err}[f(x), \phi(v)]
\end{equation}
with an error function $\mathrm{err} \in [0, 1]$ (cf., section \ref{sec:instantiation}). 

\subsection{Joint optimization}
We propose two novel strategies to jointly optimize the user and system objectives.

\paragraph{Combined sampling.}
Our first strategy 
\begin{equation}
  s_\mathrm{comb}(v) = \arg\max_{x \in \mathcal{X}} \: U(x) \: A(x, v)
\end{equation}
combines uncertainty sampling and user-oriented sampling by preferring appropriate instances for user $v$ (as in $s_\mathrm{usr}$), but among them returns the one the system is most uncertain about (as in $s_\mathrm{unc}$).

\paragraph{Trade-off sampling.}
For our second strategy, we aggregate both objectives into a single function
\begin{align}
  \hspace*{-.1cm} s_\mathrm{tos}(v) = \arg\max_{x \in \mathcal{X}} \: \big\{ \, &   (1 - \lambda) \;\: A(x, v) \\
  + & \: \lambda\; U(x) \big\} \notag
\end{align}
which is the weighted sum of user-oriented and uncertainty sampling.
The weight parameter $\lambda \in [0,1]$ can be used to adjust the learning towards the system objective or the user objective.

\section{Instantiation}
\label{sec:instantiation}
We consider our jointly optimized active learning particularly beneficial for educational applications, since (1) the users of such a system may fail to achieve their learning goals with inappropriate exercises.
Additionally, (2) it is difficult to acquire large difficulty-annotated datasets for training, as actual users are required for producing realistic training data and existing learner datasets can hardly be shared due to privacy concerns.
We therefore instantiate our approach for a language learning platform that predicts the difficulty of exercises and learns to provide appropriate (neither too easy nor too hard) exercises to its users.

\paragraph{C-tests.}
For our experiments, we use the setup of the C-test difficulty prediction task as investigated by \citet{Beinborn2016}.
C-tests are gap filling exercises proposed by \citet{Klein1982}.
In their proposed gap scheme, every second word is turned into a gap by removing the latter half of its characters. 
In contrast to cloze tests, C-tests do not require any distractors, since the first half of the word remains as a hint. 
Solving C-tests requires orthographic, morphologic, syntactic, and semantic competencies as well as general vocabulary knowledge \citep{Chapelle94}.
C-tests can be easily created automatically by choosing an arbitrary text and introducing the gaps as described above.
Because of the context and the kept word prefixes, C-test gaps typically only allow for a single solution (given by the original text) and therefore do not require manual correction.
The biggest challenge, however, lies in controlling the difficulty of the text and the derived C-test with its gaps as we have shown in previous work \citep{Lee2019}.

\begin{figure}
  \centering
  \includegraphics[width=\linewidth,trim={0 17cm 13cm 0},clip]{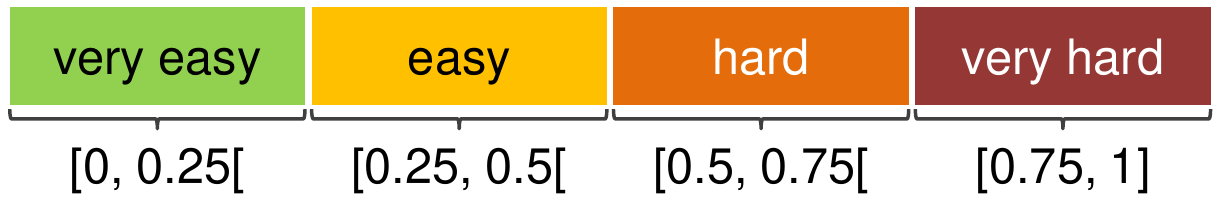}
  \caption{Gap difficulty classes and error rate ranges}
  \label{fig:gap_difficulty_range}
\end{figure}

\paragraph{System objective.}
Given a large pool $\mathcal{X}$ of C-tests $x \in \mathcal{X}$ with $n$ gaps $g_i \in x$, $1 \le i \le n$, the system objective is to learn a classifier $d(g) \in L_D$ to judge the \emph{gap difficulty} of gaps $g \in x$ with minimal training data.
As the difficulty classes $L_D$, we use the four labels \emph{very easy}, \emph{easy}, \emph{hard}, and \emph{very hard} proposed by \citet{Beinborn2016}.
These four classes are based on the mean error rates $e(g)$ of a gap $g$ observed across all users.
Figure~\ref{fig:gap_difficulty_range} shows the mapping between the mean error rates $e(g)$ and the four gap difficulty classes $L_D$.

\paragraph{Data.}
For our experiments, we obtained 3,408 solutions to English C-tests from our university's language center.
Each participant solved five C-tests with 20 gaps each (i.e., 100 gaps per solution).
The five C-tests vary across the participants based on a set of 74 different C-tests in total.
We filter out answers from 22 participants who either did not provide any correct answer or only filled out the first of the five C-tests.
Based on this dataset, we derive the ground-truth labels for the gap difficulty classification $d(g)$ based on figure~\ref{fig:gap_difficulty_range}.

\paragraph{Aggregated instances.}
In contrast to \citeauthor{Beinborn2016}'s (\citeyear{Beinborn2016}) work, a particular challenge of our setup is the need to \emph{aggregate instances}.
The active learning strategies $s(v)$ always sample entire C-tests $x \in \mathcal{X}$ and judge their appropriateness for a user $v$ based on $A(x,v)$.
The underlying classifier $d(g)$, however, operates at the level of gaps $g \in x$ within a C-test.
Similarly complex setups can be found in multiple other real-world tasks, including educational applications (e.g., providing reading recommendations at book or chapter level, but estimating appropriateness at word or sentence level) and product recommendation tasks (e.g., training a classifier for cast, plot, and action aspects, but recommending entire movies).

For our instantiation, we measure the classifier's uncertainty using the Shannon entropy
\begin{equation}
  H(g) = - \sum_{\ell \in L_D} P(\ell \mid g) \log P(\ell \mid g)
\end{equation}
across the four difficulty classes $L_D$ of a gap $g$.
$P(\ell \mid g)$ denotes the probability of the classifier $d$ to assign the difficulty class $\ell$ to gap $g$.
We then aggregate the resulting scores similar to the \emph{total token entropy} proposed by \citet{Settles2008}:
\begin{equation}
  U_\mathrm{ent}(x) = \frac{1}{n} \sum_{i = 1}^n \frac{H(g_i)}{H_\mathrm{max}}
\end{equation}
where $H_\mathrm{max}$ is the maximum achievable Shannon entropy, which serves as a normalization term.
$H_\mathrm{max}$ can be pre-computed as:
\begin{equation}\label{uent}
    H_\mathrm{max} = - \sum_{i = 1}^{|L_D|} \frac{1}{|L_D|} \log \frac{1}{|L_D|} 
\end{equation}

\paragraph{User objective.}
To model the demands of the users, we define five \emph{proficiency levels} $L_P = \{1,2,3,4,5\}$ based on the users' ability to solve C-tests.
The user representation $\phi(v) \in L_P$ of user $v$ thus returns a proficiency level between $1$ and $5$ with $5$ indicating the highest proficiency.

In our experiments, we use the C-test dataset introduced above to obtain $\phi(v)$.
Note that in this dataset, each user solved exactly five C-tests.
We therefore map their score (i.e., the percentage of correctly filled gaps) to a proficiency level that roughly corresponds to the language courses offered by the university language center.
Table~\ref{tab:proficiency_levels} shows the five levels with their corresponding score ranges and the number of users in the dataset.

\begin{table}
  \centering\small
  \begin{tabular}{l ccccc}
    \toprule
    Level & 1 & 2 & 3 & 4 & 5 \\
    \midrule
    Score (\%) & 0--54 & 55--64 & 65--74 & 75--84 & 85--100 \\
    Users & 814 & 607 & 724 & 769 & 472 \\
    \bottomrule
  \end{tabular}
  \caption{Proficiency levels, corresponding scores (\% correctly filled gaps), and number of users per level.}
  \label{tab:proficiency_levels}
\end{table}

We estimate the proficiency level of a C-test $x = g_1, g_2, \ldots ,g_n$ with
\begin{equation}
  f(x) = \psi\left(\frac{1}{n} \sum_{i=1}^{n} c(g_i)\right)
\end{equation}
where
$c\colon g \mapsto \{0, 1\}$ is an indicator function to predict if gap $g_i$ will be correctly (1) or incorrectly (0) answered and $\psi$ maps the percentage of correct answers to the corresponding proficiency level according to Table~\ref{tab:proficiency_levels}.
For our experiments, we define
\begin{equation}\label{eq-gap-prediction}
    c(g) = \left\{\begin{array}{ll}
    1 & \textrm{if } k < j \\
    0 & \textrm{otherwise} \\
  \end{array}\right. \\
\end{equation}
where $k\sim\mathcal{U}(\frac{\ell - 1}{|L_P|}, \frac{\ell}{|L_P|})$ and $j\sim\mathcal{U}(0,1)$ are uniformly sampled random variables and $\ell = d(g)$.
Based on our estimation $f(x) \in L_P$, we can now define the error function $\mathrm{err}$ as the normalized distance of $f(x)$ to the required proficiency:
\begin{equation}
  \mathrm{err}[f(x), \phi(v)] = \frac{1}{|L_P|}|f(x) - \phi(v)|
\end{equation}

\section{Experimental Setup}
\label{sec:setup}
\paragraph{System setup.}
We initialize our system with an empty set of labeled instances.
In every iteration, we sample a C-test consisting of 20 gaps from the pool of unlabeled instances $\mathcal{X}$ using one of the sampling strategies introduced in the previous section.
Then, we obtain labels based on how the user solved the test, which contributes (1) to the overall difficulty prediction for each gap and (2) to the representation of the current user's proficiency.

Our approach can be used with any underlying classifier $d(g)$.
In this paper, we train a multi-layer perceptron (MLP) to predict the four difficulty classes for a C-test gap.
To represent the input of the MLP, we use the 59 features previously proposed by \citet{Beinborn2016}. 
We furthermore introduce two novel features computed from BERT \citep{Devlin2019}:
We hypothesize that the masking objective of BERT which masks individual words during training is very similar to a gap filling exercise and thus, a model trained in such a way may provide useful signals for assessing the difficulty of a gap.
For each gap, we generate a sentence where only the gap is replaced by the masking token and fetch its predictions from the BERT model. 
From these predictions we take the prediction probability of the solution as the first feature and the entropy of the prediction probabilities of the top-50 predicted words as the second feature in concordance with findings by \citet{Felice2019} who show that entropy strongly correlates with the gap difficulty. 
Adding both features to the 59 features proposed by \citet{Beinborn2016} increases the accuracy of our MLP from 0.33 to 0.37.\footnote{The results are averaged across ten runs with different random initializations.}

While \citeauthor{Beinborn2016} successfully used support vector machines (SVM) in her work, we find that MLPs perform on par with SVMs (for the old and new features) and that they are more robust regarding the choice of the first sampled instance.
Moreover, in our initial experiments with little training data, SVMs and Logistic Regression classifiers were only able to predict the majority class. 

Our MLP has a single hidden layer consisting of $61$ hidden units.
We train the neural network for $250$ epochs with early stopping after $20$ epochs without any improvement and use \emph{Adam} \citep{Kingma2015} as our optimizer.
Note that our main interest is in the analysis of the novel active learning approach, which is why we do not systematically study the underlying classifier, but use a setup comparable to the state-of-the-art results reported by \citet{Beinborn2016}.

We run experiments for each of our sampling strategy. 
We select five C-tests without any overlap between users, texts, and their corresponding user answers to create an independent test set and put the remaining 69 C-tests into the pool of unlabeled data. 
In the first iteration, we use the randomly initialized weights of our neural network to select the starting example. 
To provide comparable results between different runs, we keep the parameter initialization of our neural network fixed when comparing different sampling strategies.
We limit each experimental run to $8 \cdot 5 = 40$ iterations, as the five proficiency levels are not evenly distributed with the smallest class having only eight C-tests.
At each iteration, we train our model on 80\% of the already labeled data and use the remaining 20\% as our validation set (split randomly).
We use the best-performing model on the validation set for testing and store it as our model initialization for the next iteration.
On an \textit{Intel Core i5-4590}, a single run with 40 iterations takes less than four minutes.

\paragraph{Learner behavior.}

To study the benefit of our approach for different types of learners,\footnote{Henceforth, we use \emph{learner} to refer to the users of an educational application rather than to a machine learning system.} we derive four prototypical learner behaviors from our C-test dataset.
To prepare this, we first compile a probabilistic model for the learners of each proficiency group as described in Table~\ref{tab:proficiency_levels} to obtain learner-specific gap error rates $e(g,v)$.
The learner-specific gap error rates are computed by binning all learners into the specific groups and then computing the error rate by averaging for each gap.
If there is no error rate for a given gap and learner in our dataset, we use the averaged gap error rate of the corresponding proficiency group to simulate an answer.

Using these learner-specific gap error rates, we  predict whether an answer to a  C-test gap $g$ is correct or incorrect similar to Equation~(\ref{eq-gap-prediction}):
\begin{equation}
  \hat{c}(g) = \left\{\begin{array}{ll}
    1 & \textrm{if } e(g,v) < j \\
    0 & \textrm{otherwise} \\
  \end{array}\right. \\
\end{equation}
In contrast to Equation~(\ref{eq-gap-prediction}), we do not sample $k$, but use the learner-specific error rates $e(g,v)$ for gap $g_i$ from the proficiency level $\phi(v)$. Again, $j\sim\mathcal{U}(0,1)$ is a uniformly sampled random variable.

For a language learning platform, it is likely that motivated learners who continually practice improve their proficiency over time.
Less motivated learners or learners who suffer from distractions, interruptions, or frustration, however, may show different paces in their learning speed or even deteriorate in their proficiency.  
Therefore, we study four prototypical types of learner behavior:

\begin{itemize}[noitemsep,topsep=3pt,itemsep=3pt,itemindent=-1em]
\renewcommand\labelitemi{--}
\item
\emph{Static learners} ($\mathrm{STAT}$) do not improve their skills over the course of our experiments. Instead, they provide answers constantly at the same, pre-defined proficiency level.
This models learners with a slow progress or with little motivation overall.

\item
\emph{Motivated learners} ($\mathrm{MOT}$) continually improve their language proficiency throughout our experiments with a fixed step size of $t_1$ C-tests.
That is, we simulate that their proficiency level $\phi(v)$ increases by one every $t_1$ iterations.

\item
\emph{Interrupted learners} ($\mathrm{INT}$) experience a drop in their proficiency during our experiments. 
Such cases occur, for example, if a learner has to interrupt their learning process for a longer time.
For our simulation, we start with the motivated learner setup, constantly increasing the proficiency every $t_1$ iterations.
However, this learner experiences a sudden increase ($t_2$) and drop ($t_3$) in the proficiency level by one.
After recovering from the drop ($t_4$) the proficiency will again increase according to the motivated learner ($t_5$).

\item \emph{Artificially decreasing learner.} ($\mathrm{DEC}$) Finally, our last group of simulated learners displays a constant drop in their proficiency during our simulation. 
Although such cases rarely occur in the real world, we use this learner to evaluate all sampling strategies in the case of constant drop. 
Similar to the motivated learner, we start with the highest possible proficiency and decrease it by one every $t_1$ iterations.

\end{itemize}

\noindent
For our experiments, we assume a static learner that remains at proficiency level $\phi(v)=3$.
For motivated learners, we set the initial proficiency level to $1$ and use a step size of $t_1 = 8$, so that they traverse all proficiency levels throughout a single run.
For interrupted learners, we also use $t_1 = 8$ with an additional increase after $t_2=12$, a drop after $t_3=16$, and a recovery (increase) after $t_4=20$. 
Starting from $t_5=24$, interrupted learners behave the same as motivated learners.

Like \citet{Beinborn2016}, we cannot publish the C-test data due to data privacy reasons, but we provide our code and simulated learner models on GitHub.\footnote{\url{https://github.com/UKPLab/acl2020-empowering-active-learning}}
\section{Experiments}
\label{sec:results}
We present and discuss our results for $U_\mathrm{ent}$ and $A$ as defined in section~\ref{sec:instantiation}.
For each strategy we run our experiments ten times with different weight initializations and report the averaged scores.
For random sampling, we do ten runs with different random seeds for each weight initialization to provide more stable results.
We set $\lambda=0.5$ for our trade-off sampling strategy.

\subsection{Evaluation metrics}
As our system and user objectives have different scopes (gap-level vs.\ exercise-level), we quantify both differently.  
To measure the system objective, we report the \textit{accuracy} of our model for predicting the individual gap difficulties of the test data after each iteration. 
As our training data increases by 20 gaps after each iteration, we provide plots for all experiments from the first to the last (40-th) iteration. 
For quantifying the user objective, we evaluate all sampling strategies across all 40 iterations, i.e., how well our sampling strategies were able to satisfy the user's needs after the whole set of exercises.
Instead of accuracy, we take the distance-based metric \textit{mean absolute error} ($\mathrm{MAE}$). 
As users explicitly query a C-test of a specific proficiency level at each iteration, suggesting a C-test which deviates by two levels from the requested proficiency has a worse impact on the user's learning experience than a C-test which only deviates by one level.
For better interpretability, we do not normalize the $\mathrm{MAE}$ as we do for our error function $\mathrm{err}$, i.e., a $\mathrm{MAE}$ of 1 means that on average, the difficulty of the sampled instances was off by a whole proficiency level from the queried ones.

\subsection{Results}
Since the interrupted learner experiences both a drop and increase in proficiency in a less constant manner than the motivated or decreasing learners, we conduct further analysis of our sampling strategies for the interrupted learner.

\paragraph{System objective.}

\begin{figure}
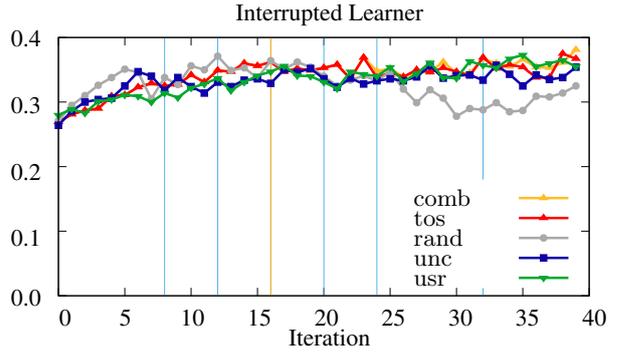

\centering
    \include{plots/interrupted_tokenentropy}
  \caption{Accuracy on the test data for $U_\mathrm{ent}$.} 
  \label{fig:accuracy_uent}
\end{figure}

Figure \ref{fig:accuracy_uent} shows the system objective for $U_\mathrm{ent}$ after each iteration.
Vertical blue lines indicate increases in the learner's proficiency whereas the vertical yellow line indicates a drop.
We observe that although random sampling performs rather well in the early iterations, all our proposed strategies as well as the uncertainty sampling baseline are able to outperform it in the later iterations.
Moreover, all proposed strategies perform similar to uncertainty sampling.
This is surprising, especially for the user-oriented sampling strategy as it inherently does not optimize the system objective.
One reason for this may be the similarity of the user-oriented sampling strategy to \textit{curriculum learning} \citep{Bengio2009}, which opts to organize model training in a meaningful way.
As we sample instances the model is most confident in (i.e., have the highest prediction confidence) this leads to instances which are easier to learn and may especially be helpful in low-data scenarios.

To better quantify our results, we compare the averaged accuracy scores across all iterations, shown in table~\ref{tab:mean-u-ent} and conduct Wilcoxon signed-rank tests \citep{Wilcoxon1992} on the active learning curves for system and model objectives to test for statistical significance.
We can observe that for the static, motivated, and interrupted learners both our joint sampling strategies outperform all baselines significantly (p $<$ 0.05), but show no significant difference between each other.\footnote{The system performance of random sampling remains the same for all learner types as it is averaged across all runs.}
Only for the decreasing learner all strategies show no significant difference at all.
In concordance with our observations for the user-oriented sampling which may benefit from first sampling easy-to-learn instances, jointly optimizing system and user objective seems to benefit from curriculum learning and active learning paradigms.

\begin{table}
  \centering
  \small{
  \begin{tabular}{lcccc}
    \toprule
    {} & $\mathrm{STAT}$ & $\mathrm{MOT}$ & $\mathrm{INT}$ & $\mathrm{DEC}$\\
    \midrule
    {$\mathrm{tos}$} &   \textbf{.344} & .338 & .339 & .327 \\
    {$\mathrm{comb}$ } &   .343 & \textbf{.340} & \textbf{.341} & .327 \\
    {$\mathrm{usr}$}      &   .338 & .331 & .334 & .328 \\
    {$\mathrm{unc}$} & .332 & .331 & .331 & \textbf{.331} \\
    {$\mathrm{rand}$}     & .325 & .325 & .325 & .325 \\
    \bottomrule
  \end{tabular}
  \caption{Averaged accuracy over all iterations for $U_\mathrm{ent}$}
  \label{tab:mean-u-ent}
  }
\end{table}

\paragraph{User objective.}

\begin{table}
  \centering
  \small{
  \begin{tabular}{lcccc}
    \toprule
    {} & $\mathrm{STAT}$ & $\mathrm{MOT}$ & $\mathrm{INT}$ & $\mathrm{DEC}$\\
    \midrule
    {$\mathrm{tos}$} & 0.98 & 0.65 & 0.93 & 0.75 \\
    {$\mathrm{comb}$ } & 0.98 & 0.63 & 0.88 & \textbf{0.65} \\
    {$\mathrm{usr}$} & \textbf{0.85} & \textbf{0.58} & \textbf{0.65} & 0.75 \\
    {$\mathrm{unc}$} & 1.17 & 1.33 & 1.35 & 1.72 \\
    {$\mathrm{rand}$ } & 1.16 & 1.22 & 1.82 & 1.24  \\
    \bottomrule
  \end{tabular}
  \caption{$\mathrm{MAE}$ for $U_\mathrm{ent}$}
  \label{tab:a-u-comparison-unc_uent}
  }
\end{table}

Table \ref{tab:a-u-comparison-unc_uent} shows the $\mathrm{MAE}$ for all strategies using $U_\mathrm{ent}$. 
We can observe that all strategies which consider a separate user objective sample instances which significantly better fit the current user proficiency.\footnote{Statistical testing was again conducted using a Wilcoxon signed-rank test for p $<$ 0.05.}
Furthermore, the combined sampling approach which puts more emphasis on the user objective outperforms our trade-off sampling for all learner behaviors and even manages to outperform the user-oriented sampling strategy for the decreasing learner.

\begin{figure}
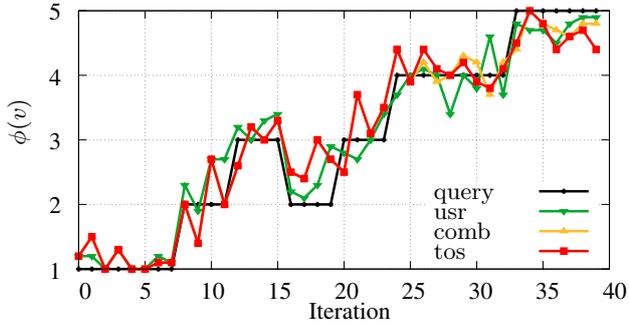

    \centering
      \include{plots/interrupted_tokenentropy_user}
    \caption{Sampled instances for the interrupted learner.}
    \label{fig:user_objective-uent}
\end{figure}

We further investigate how well our approaches react to changes in the user objective by plotting the mean difficulty $\phi(v)$ of sampled instances after each step for all our strategies modeling the user objective.
As figure \ref{fig:user_objective-uent} shows, all sampling strategies are able to match the queried C-test difficulties well, as they do not deviate much from the queried difficulty (in black).

\paragraph{Adaptive choice of $\lambda$.} We furthermore investigate how the choice of $\lambda$ affects our trade-off sampling strategy.
As the system predictions may not be very accurate in early iterations, it is reasonable to put more emphasis on the system objective in the beginning, but focus on providing suited C-tests (user objective) in later iterations.
We thus define $\lambda$ as an adaptive function $\lambda = f(i)=\frac{1}{\sqrt{i}}=i^{-0.5}$ which highly emphasizes the system objective in early stages and anneals with an increasing number of iterations $i$.

\begin{figure}
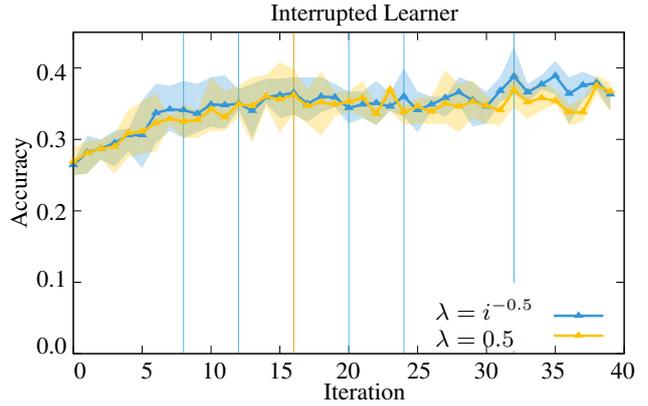

    \centering
      \include{plots/interrupted_tokenentropy_lambda}
    \caption{Accuracy of $\mathrm{tos}$ for annealed and fixed $\lambda$.}
    \label{fig:lambda-uent}
\end{figure}

Figure~\ref{fig:lambda-uent} shows the system performance of our trade-off sampling strategy averaged across ten different runs.
The colored areas show the corresponding upper and lower quartiles. 
As shown in table~\ref{tab:adaptive-lambda}, we can see that our annealed $\lambda$ leads to considerable improvements for system and user objective, leading to a significant increase in average accuracy from $0.339$ to $0.347$ and a decrease in the $\mathrm{MAE}$ from $0.93$ to $0.48$ for the interrupted learner, outperforming all other sampling strategies. 

\begin{table}
  \small
  \centering
  \begin{tabular}{l*{4}c}
    \toprule
    Acc & $\mathrm{STAT}$ & $\mathrm{MOT}$ & $\mathrm{INT}$ & $\mathrm{DEC}$  \\
    \midrule
    $\mathrm{tos}_\lambda$ & .333 & \textbf{.346} & \textbf{.347} & .314 \\
    $\mathrm{tos}$ & .334 & .338 & .339 & .327 \\
    \midrule
    MAE &  $\mathrm{STAT}$ & $\mathrm{MOT}$ & $\mathrm{INT}$ & $\mathrm{DEC}$  \\
    \midrule
    $\mathrm{tos}_\lambda$ & \textbf{0.85} & \textbf{0.53} & \textbf{0.48} & \textbf{0.53}  \\
    $\mathrm{tos}$ & 0.98 & 0.65 & 0.93 & 0.75 \\
    \bottomrule
  \end{tabular}
  \caption{Averaged accuracy scores and $\mathrm{MAE}$ with an annealed $\lambda$ for $U_\mathrm{ent}$.}
  \label{tab:adaptive-lambda}
 \end{table}

\paragraph{Further findings.}
We observe similar results for system and user objectives for the other learner types. 
Investigating the stability of all sampling approaches furthermore shows that our joint optimization strategies perform better and more stable in early iterations.

Due to averaging, $U_\mathrm{ent}$ cannot distinguish between C-tests with only a few highly uncertain gaps and C-tests which have a higher number of less uncertain gaps.
However, in preliminary experiments with a different aggregation function which is more robust to C-tests with only a few highly uncertain gaps, we come to similar findings across all sampling strategies and learner types. 
Detailed results for our other learner behaviors, the stability of our sampling strategies, and the results of our preliminary experiments with a different aggregation function are provided in the paper's appendix.

\paragraph{Limitations.}
Although our setup with simulated learners may seem artificial compared to an evaluation study with real-world learners, to conduct such a study in an ethical way, we need to ensure that participants are not hurt in their learning process.
Thus, strategies which can be evaluated in user studies are limited to those which consider the user objective.
In contrast, the use of simulated learners allows us to compare our proposed strategies against common active learning strategies which do not consider the user objective at all.

Another limitation is how to estimate a learner's current proficiency given that we do not know the true difficulty of a C-test. 
This raises the general question of using relative or absolute difficulties for the selection of suited exercises. 
In this work, we assumed absolute proficiency levels and implemented according learner behaviors to provide a more controlled environment for our experiments. 
In the case of absence of any absolute (true) difficulty estimations for C-tests, we see several directions for future work:

\begin{enumerate}[a)]
    \item As a simple baseline, a normalized version of $\psi(x)$ may be applied on a learner's previously filled-out C-tests.
    However, this assumes that all C-tests are equally difficult which may lead to unsuited C-tests. 
    \item Training an additional model for assessing a learner's proficiency given their results on a C-test with the gap-difficulty predictions from our model serving as additional input.
    \item Instead of using the absolute difficulty, one may define an optimal error margin as a zone of proximal development \citep{Vygotsky78}.
    This requires an adaptation of the user objective to the relative difficulties of exercises for individual learners, but may be an important step in achieving highly personalized user models without any absolute labels. 
\end{enumerate}

\section{Conclusion}
\label{sec:conclusion}
In this work, we investigated how we can incorporate user feedback into existing active learning approaches without hurting the user's actual needs.
We formalize both \textit{system} (active learning) and \textit{user} objectives and propose two novel sampling strategies which aim to maximize both objectives jointly.
We evaluate our sampling strategies for the task of selecting suited C-tests, a type of fill-the-gap exercise, which fit the current proficiency of a human learner.
We create simulated learners for five different proficiency levels from real-world data and use them to define different learning behaviors.
Our experiments show that both our novel sampling strategies are successfully selecting instances which lead to a better model training while not hurting a learner's progress by selecting too easy or too difficult C-tests.
Although system and user objective at first seem counteracting, our experiments indicate that they complement each other as jointly optimizing them outperforms optimizing only one of the goals.
Additional experiments with an adaptive $\lambda$ for our trade-off sampling strategy show that properly balancing system and user objective can lead to considerable improvements in performance for both objectives.

Our findings open up new opportunities for training models on low-resource scenarios with implicitly collected user feedback while jointly serving the user's actual needs.
Additional use cases like the training of personalized recommendation models as well as the use of reinforcement learning to find a good trade-off between system and user objective remain to be investigated in future work.

\section*{Acknowledgments}
This work has been supported by the German Research Foundation with the ArguAna project (GU~798/20-1) and the Evidence project (GU~798/27-1).
We thank the anonymous reviewers for their detailed and helpful comments as well as Edwin Simpson and Yevgeniy Puzikov for the insightful discussions about our work.
We especially thank the language center of the Technische Universit\"at Darmstadt for providing us with the data and Dr.\ Lisa Beinborn for providing us with the code to extract her proposed features.

\bibliography{acl2020}
\bibliographystyle{acl_natbib}

\appendix
\section{Appendices}
\label{sec:appendix}
\subsection{Results of $U_\mathrm{ent}$ for other learner types}
\label{sec:appendix-uent-results}

\begin{figure*}[!htb]
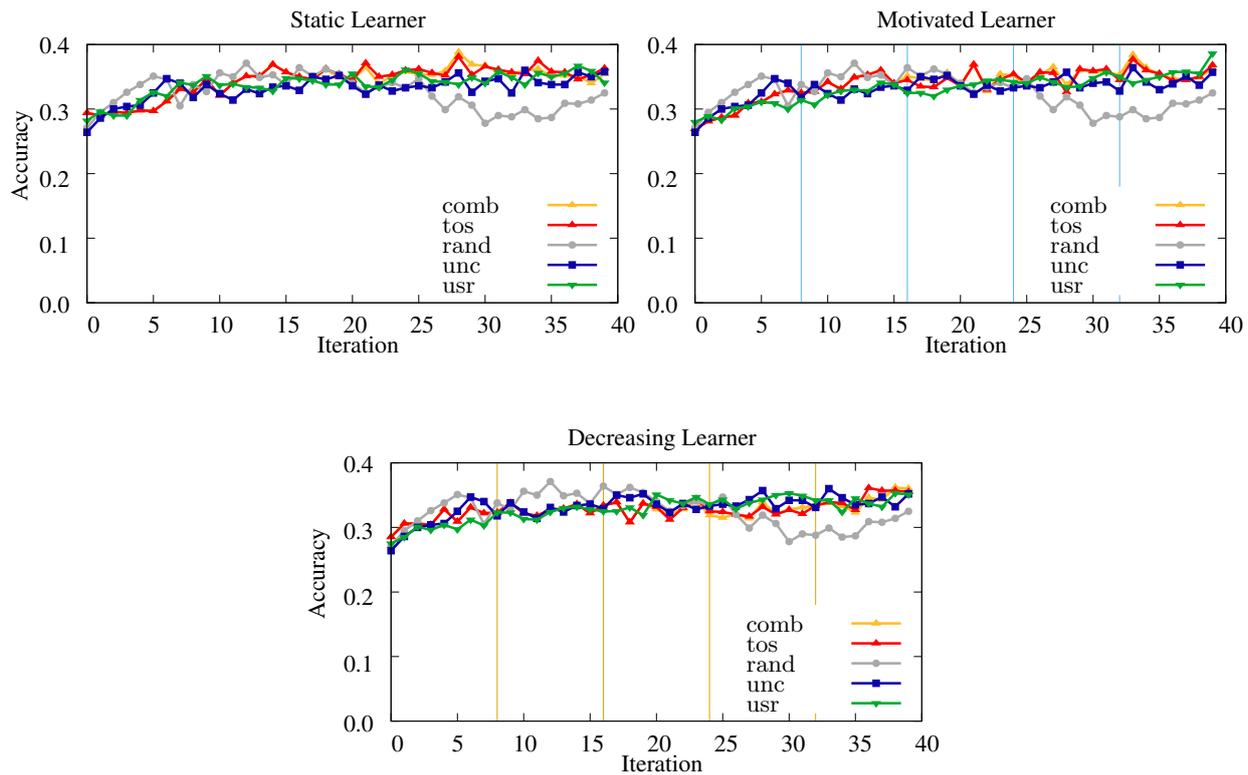

    \centering
    \begin{minipage}{0.5\textwidth}
        \centering
          \include{plots/static_tokenentropy}
    \end{minipage}\hfill
    \begin{minipage}{0.5\textwidth}
        \centering
          \include{plots/increasing_tokenentropy}
    \end{minipage}
\\
    \begin{minipage}{0.5\textwidth}
        \centering
          \include{plots/decreasing_tokenentropy}
    \end{minipage}
  \caption{Accuracy scores for the static, motivated, and artificially decreasing learners using $U_\mathrm{ent}$.}
    \label{fig:u-ent-other-learners}
\end{figure*}

 Figure~\ref{fig:u-ent-other-learners} shows our results for the static, motivated, and artificially decreasing learner.
 As with the interrupted learner, blue (yellow) vertical lines indicate an increase (drop) in the learner's proficiency. 
 Similar to the results for the interrupted learner, all strategies outperform random sampling in later iterations.

%%%%%%%%%%%%%%%%%%%%%%%%%%%%%%%%%%%%%%%%%%%%
%          U-softmax formulation
%%%%%%%%%%%%%%%%%%%%%%%%%%%%%%%%%%%%%%%%%%%%

\subsection{An outlier-invariant variation of $U$}
\label{sec:appendix-usoft}
Due to averaging, $U_\mathrm{ent}$ cannot distinguish between C-tests with only a few highly uncertain gaps and C-tests which have a higher number of less uncertain gaps.
We investigated another aggregation function $U_\mathrm{soft}$ in preliminary experiments, which measures the entropy across all gaps and thus, is more robust to C-tests with only a few highly uncertain gaps.

\paragraph{Formulation.}
For our second formulation of $U$, we use a different aggregation method. 
Due to the mean, $U_\mathrm{ent}$ is unable to distinguish between C-tests where the system is highly uncertain for only a few gaps and C-tests where all gaps are less, but more equally uncertain.
We propose to use the softmax function $\sigma$ for normalizing $H(g_i)$ and then to compute the entropy across all gaps $g_i$.
$U_\mathrm{soft}$ thus considers the distribution of gap-uncertainties and favours C-tests with equally distributed gap-uncertainties over C-tests with only a few highly uncertain gaps.
\begin{equation}
    U_\mathrm{soft}(x) = \gamma \ [-\sum_{i=1}^{n} \sigma_i(H(g_i))\ \mathrm{log} \ \sigma_i(H(g_i))]
\end{equation}
As the squashing of the individual gap entropy values removes the information about their magnitude, we furthermore scale the resulting value by the normalized mean entropy 
\begin{equation}
\gamma = \frac{1}{ n \log n} \sum_{i=1}^{n} \frac{H(g_i)}{H_{\mathrm{max}}}
\end{equation}
for all gaps $g_i$ in the C-test.

%%%%%%%%%%%%%%%%%%%%%%%%%%%%%%%%%%%%%%%%%%%%
%       RESULTS
%%%%%%%%%%%%%%%%%%%%%%%%%%%%%%%%%%%%%%%%%%%%

\paragraph{Results.}

\begin{figure*}[!htb]
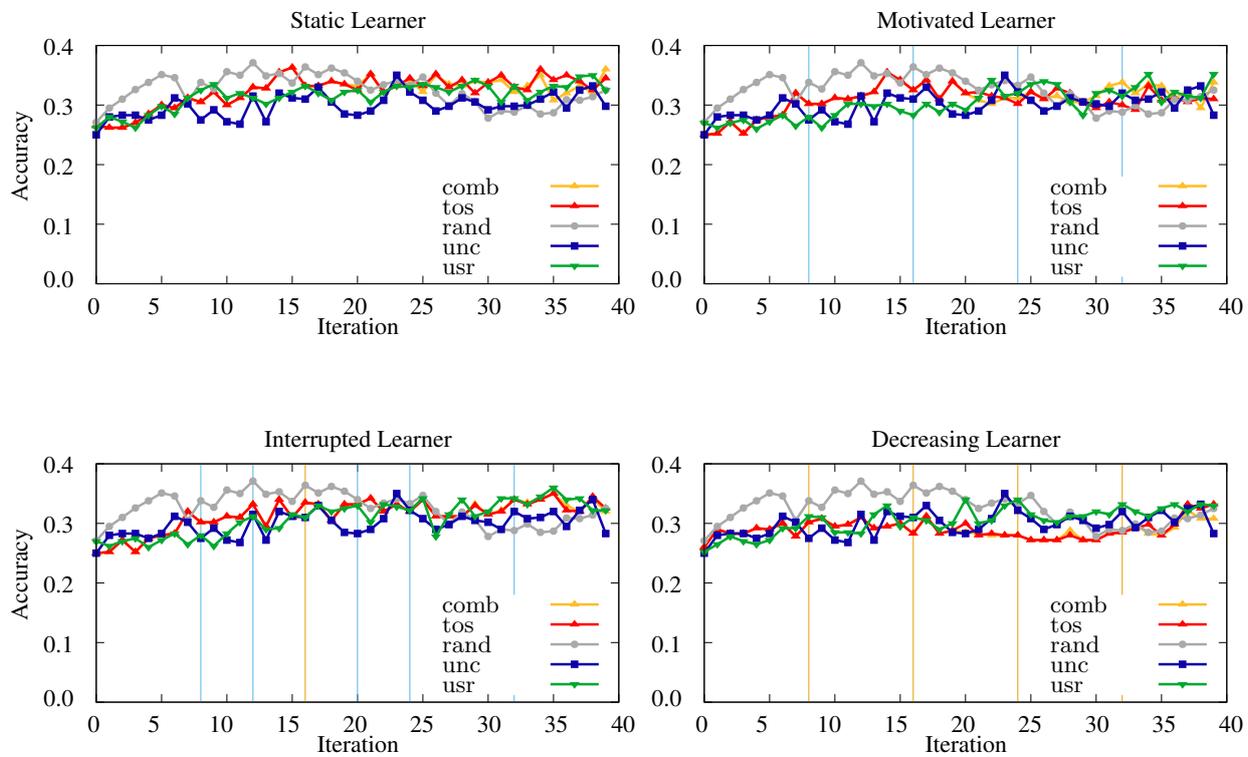

\centering
    \begin{minipage}{0.5\textwidth}
        \centering
          \include{plots/static_softentropy}
    \end{minipage}\hfill
    \begin{minipage}{0.5\textwidth}
        \centering
          \include{plots/increasing_softentropy}
    \end{minipage}
    \\
    \begin{minipage}{0.5\textwidth}
        \centering
          \include{plots/interrupted_softentropy}
    \end{minipage}\hfill
    \begin{minipage}{0.5\textwidth}
        \centering
          \include{plots/decreasing_softentropy}
    \end{minipage}
  \caption{Accuracy on the test data for $U_\mathrm{soft}$.}
  \label{fig:accuracy_usoft}
\end{figure*}

Figure~\ref{fig:accuracy_usoft-comparisons} shows similar tendencies as we already found for $U_\mathrm{ent}$ in section~6.
Again, we can observe that random sampling performs better in early iterations, while the other sampling strategies outperform it in latter iterations.
Averaging the accuracy across all iterations (table~\ref{tab:all-results-with-lambdas}) shows that both our joint sampling strategies $\mathrm{tos}$ and $\mathrm{comb}$ again perform in average better than the other sampling strategies for the static, motivated, and interrupted learners.
However, conducting a Wilcoxon signed-rank test with $p < 0.05$ shows that the active learning curves only significantly differ for the static learner.

\begin{figure}[H]
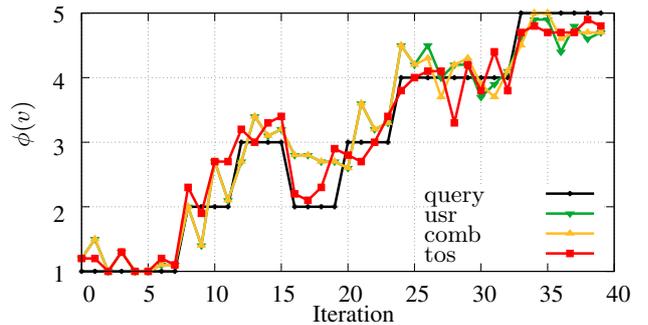

    \centering
      \include{plots/interrupted_softentropy_user}
    \caption{Sampled instances for the interrupted learner using $U_\mathrm{soft}$.}
    \label{fig:user_objective-usoft}
\end{figure}

For the user objective (also shown in table~\ref{tab:all-results-with-lambdas}) we observe that all strategies which include a user objective significantly outperform $\mathrm{rand}$ and $\mathrm{unc}$, but there is no clear favorite amongst them.
This can also be seen in figure~\ref{fig:user_objective-usoft} where all strategies manage to sample instances close to the queried difficulty (in black).

 \subsection{Impact of the aggregation function}
\label{sec:appendix-uent-vs-usoft}
\begin{figure}[!htb]
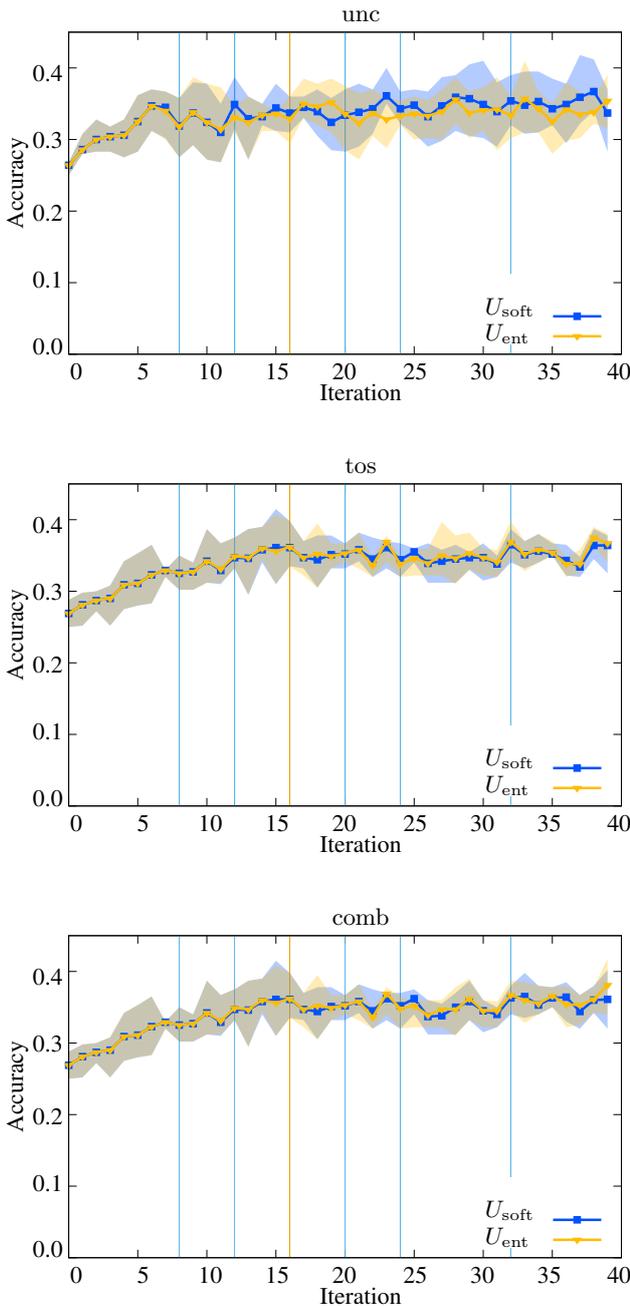

\centering
    \include{plots/interrupted_soft-vs-ent_uncertainty}
    
    \include{plots/interrupted_soft-vs-ent_tradeoff}
    
    \include{plots/interrupted_soft-vs-ent_combined}
  \caption{Comparing $U_\mathrm{ent}$ and $U_\mathrm{soft}$ for the interrupted learner.} 
  \label{fig:accuracy_uent-usoft}
\end{figure}

Figure \ref{fig:accuracy_uent-usoft} compares both our aggregation functions $U_\mathrm{ent}$ and $U_\mathrm{soft}$ against each other on the interrupted learner for uncertainty, combined, and trade-off sampling.
Although $U_\mathrm{ent}$ and $U_\mathrm{soft}$ differ to some regard, directly comparing both aggregation functions and the respective aggregated scores (cf., table~\ref{tab:all-results-with-lambdas} shows that there is no clear favourite between both.
Extensive work with respect to both aggregation functions as well as additional aggregation strategies remains to be investigated in future work.

 \subsection{Stability of system objective}
 \label{sec:appendix-stability}
\begin{figure*}[!htb]
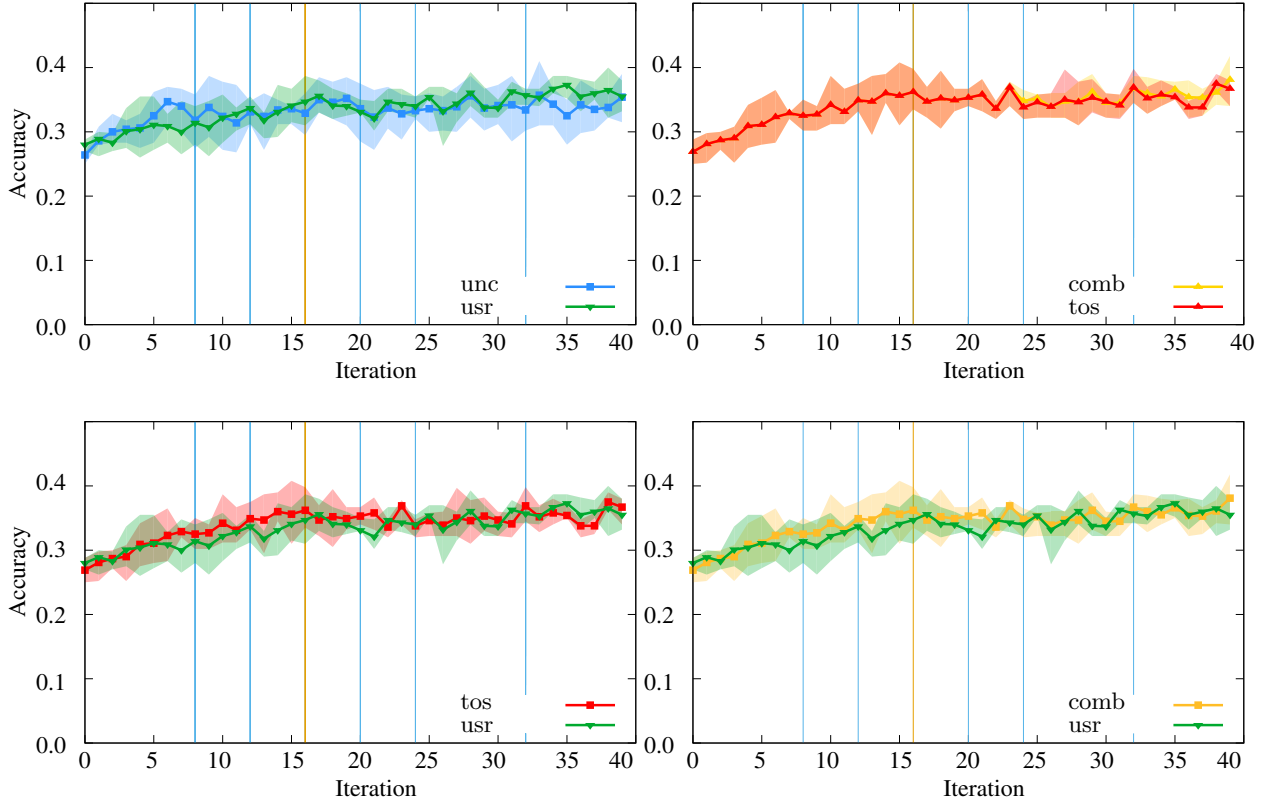

\centering
    \begin{minipage}{0.5\textwidth}
        \centering
          \include{plots/interrupted_tokenentropy_percentile_uncertainty-user}
    \end{minipage}\hfill
    \begin{minipage}{0.5\textwidth}
        \centering
          \include{plots/interrupted_tokenentropy_percentile_tradeoff-combined}
    \end{minipage}
    \\
    \begin{minipage}{0.5\textwidth}
        \centering
          \include{plots/interrupted_tokenentropy_percentile_tradeoff-user}
    \end{minipage}\hfill
    \begin{minipage}{0.5\textwidth}
        \centering
          \include{plots/interrupted_tokenentropy_percentile_combined-user}
    \end{minipage}
  \caption{Upper and lower quartiles for the interrupted learner using $U_\mathrm{ent}$.} 
  \label{fig:accuracy_uent-comparisons}
\end{figure*}

 To provide estimates how stable our approaches are across different randomly initialized weights, we compute the upper and lower quartiles for each sampling strategy across all runs.
Figures~\ref{fig:accuracy_uent-comparisons} and \ref{fig:accuracy_usoft-comparisons} show our results for the interrupted learner.

Overall, we observe that user-oriented sampling has lower deviations across different runs for both our aggregation functions $U_\mathrm{ent}$ and $U_\mathrm{soft}$. 
One reason for this may be that in contrast to uncertainty sampling, we query instances with highly certain predictions in our user-oriented sampling approach. 
This leads to sampled instances which are easier to learn resulting in a higher training stability with small data.
Comparing the user-oriented against our joint sampling strategies shows that especially in the earlier iterations, our proposed sampling strategies perform better and provide more stable training. 

\begin{figure*}
\centering
    \begin{minipage}{0.5\textwidth}
        \centering
          \include{plots/interrupted_softentropy_uncertainty-user}
    \end{minipage}\hfill
    \begin{minipage}{0.5\textwidth}
        \centering
          \include{plots/interrupted_softentropy_tradeoff-combined}
    \end{minipage}
    \\ 
    \begin{minipage}{0.5\textwidth}
        \centering
          \include{plots/interrupted_softentropy_tradeoff-user}
    \end{minipage}\hfill
    \begin{minipage}{0.5\textwidth}
        \centering
          \include{plots/interrupted_softentropy_combined-user}
    \end{minipage}
  \caption{Upper and lower quartiles for the interrupted learner using $U_\mathrm{soft}$.} 
  \label{fig:accuracy_usoft-comparisons}
\end{figure*}

\subsection{Further investigation of $\lambda$}

\begin{figure}[H]
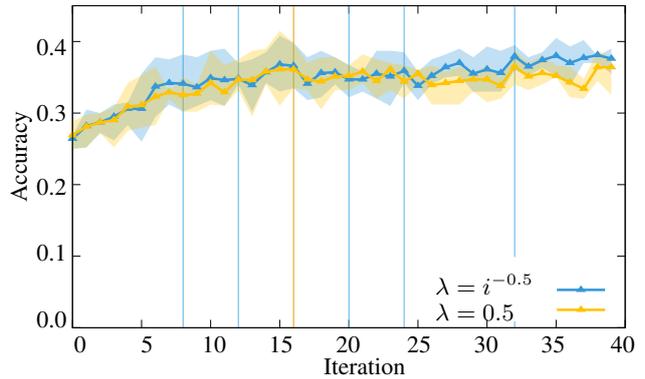

    \centering
      \include{plots/interrupted_softentropy_lambda}
    \caption{Accuracy of trade-off sampling for annealed and fixed $\lambda$ using $U_\mathrm{soft}$ for the interrupted learner.}
    \label{fig:lambda-usoft}
\end{figure}

To further validate our findings for an annealed $\lambda$, we conduct the same experiments with our novel aggregation function $U_\mathrm{soft}$.
As with $U_\mathrm{ent}$, we obtain significant improvements for our trade-off sampling strategy (figure~\ref{fig:lambda-usoft}) for the motivated and interrupted learner, but also a significant decrease for the static and decreasing learner.
With respect to the user objective, we do not see any significant differences at all, indicating that $U_\mathrm{soft}$ does not benefit at all from the emphasised user objective in later iterations.

\begin{table*}
  \small
  \begin{tabular}{l *{4}{@{\hspace{.1cm}}*{4}{@{\hspace{.2cm}}c}}}
    \toprule
    {} & \multicolumn{8}{c}{$\mathrm{U_\mathrm{ent}}$} & \multicolumn{8}{c}{$\mathrm{U_\mathrm{soft}}$} \\
    \cmidrule(lr){2-9} \cmidrule(lr){10-17} 
    {} & \multicolumn{4}{c}{Accuracy} & \multicolumn{4}{c}{$\mathrm{MAE}$} & \multicolumn{4}{c}{Accuracy} & \multicolumn{4}{c}{$\mathrm{MAE}$} \\
    \cmidrule(lr){2-5} \cmidrule(lr){6-9} 
    \cmidrule(lr){10-13} \cmidrule(lr){14-17} 
    %$s$ & 
     & $\mathrm{STAT}$ & $\mathrm{MOT}$ & $\mathrm{INT}$ & $\mathrm{DEC}$ & 
     $\mathrm{STAT}$ & $\mathrm{MOT}$ & $\mathrm{INT}$ & $\mathrm{DEC}$ & 
     $\mathrm{STAT}$ & $\mathrm{MOT}$ & $\mathrm{INT}$ & $\mathrm{DEC}$ & 
     $\mathrm{STAT}$ & $\mathrm{MOT}$ & $\mathrm{INT}$ & $\mathrm{DEC}$ \\
    \midrule
    $\mathrm{tos}_\lambda$ & 
    .333 & .346 & .347 & .314 & 
    0.85 & 0.53 & 0.48 & 0.53 &
    .331 & .345 & .347 & .316 &
    0.86 & 0.64 & 0.59 & 0.70 \\
    $\mathrm{tos}$ & 
    .334 & .338 & .339 & .327 & 
    0.98 & 0.65 & 0.93 & 0.75 &
    .345 & .336 & .338 & .327 &
    0.91 & 0.64 & 0.62 & 0.70 \\
    $\mathrm{comb}$ & 
    .343 & .340 & .341 & .327 & 
    0.98 & 0.63 & 0.88 & 0.65 &
    .344 & .338 & .340 & .326 &
    0.93 & 0.63 & 0.62 & 0.66 \\
    $\mathrm{usr}$ & 
    .338 & .331 & .334 & .328 &
    0.85 & 0.58 & 0.65 & 0.75 &
    .337 & .331 & .334 & .328 &
    0.92 & 0.63 & 0.59 & 0.70 \\
    $\mathrm{unc}$ & 
    .332 & .331 & .331 & .331 &
    1.17 & 1.33 & 1.35 & 1.72 &
    .336 & .336 & .336 & .335 &
    1.24 & 1.32 & 1.31 & 1.72 \\

    $\mathrm{rand}$ &
    .325 & .325 & .325 & .325 &
    1.16 & 1.22 & 1.82 & 1.24 &
    .325 & .325 & .325 & .325 &
    1.16 & 1.22 & 1.82 & 1.24 \\
    \bottomrule
  \end{tabular}
  \caption{Averaged accuracy and $\mathrm{MAE}$ for all strategies (including the annealed $\lambda$ strategy) for $U_\mathrm{ent}$ and $U_\mathrm{soft}$.}
  \label{tab:all-results-with-lambdas}
 \end{table*}

Table~\ref{tab:all-results-with-lambdas} (including the previous results for better comparability) shows the results for all learner behaviours and both our aggregation functions $U_\mathrm{ent}$ and $U_\mathrm{soft}$.
As can be seen, using an annealed $\lambda$ ($\mathrm{tos}_\lambda$) leads to the best results with respect to the user objective for $U_\mathrm{ent}$ but fails to improve the results for $U_\mathrm{soft}$.

\end{document}